\documentclass{article} %
\PassOptionsToPackage{usenames,dvipsnames}{xcolor}
\usepackage{iclr2022_conference,times}

\usepackage{amsmath,amsfonts,bm}

\def\eqref#1{equation~\ref{#1}}

\def\1{\bm{1}}

\DeclareMathAlphabet{\mathsfit}{\encodingdefault}{\sfdefault}{m}{sl}
\SetMathAlphabet{\mathsfit}{bold}{\encodingdefault}{\sfdefault}{bx}{n}

\usepackage{bbm}

\usepackage{hyperref}
\usepackage{url}

\usepackage{adjustbox}

\usepackage{algorithm}
\usepackage{algpseudocode}

\usepackage{wrapfig}

\usepackage{tikz}
\usepackage{pgfplots}
\DeclareUnicodeCharacter{2212}{−}
\usepgfplotslibrary{groupplots,dateplot}
\usetikzlibrary{patterns,shapes.arrows}
\pgfplotsset{compat=newest}
\usetikzlibrary{arrows}
\usepackage{url}
\usepackage{booktabs}

\usetikzlibrary{math}
\tikzmath
{
  function symlog(\x,\a){
    \yLarge = ((\x>\a) - (\x<-\a)) * (ln(max(abs(\x/\a),1)) + 1);
    \ySmall = (\x >= -\a) * (\x <= \a) * \x / \a ;
    return \yLarge + \ySmall ;
  };
  function symexp(\y,\a){
    \xLarge = ((\y>1) - (\y<-1)) * \a * exp(abs(\y) - 1) ;
    \xSmall = (\y>=-1) * (\y<=1) * \a * \y ;
    return \xLarge + \xSmall ;
  };
}

\usepackage[pdf]{graphviz}

\usepgfplotslibrary{external} 
\title{A Siren Song of Open Source Reproducibility}

\author{Edward Raff \\
Booz Allen Hamilton\\
University of Maryland, Baltimore County \\
\texttt{raff\_edward@bah.com} \AND
Andrew L. Farris \\
Booz Allen Hamilton\\
\texttt{Farris\_Drew@bah.com} 
}

\iclrfinalcopy %
\begin{document}

\maketitle

\begin{abstract}
As reproducibility becomes a greater concern, conferences have largely converged to a strategy of asking reviewers to indicate whether code was attached to a submission. This is part of a larger trend of taking action based on assumed ideals, without studying if those actions will yield the desired outcome. Our argument is that this focus on code for replication is misguided if we want to improve the state of reproducible research. This focus can be harmful --- we should not force code to be submitted. There is a lack of evidence for effective actions taken by conferences to encourage and reward reproducibility. We argue that venues must take more action to advance reproducible machine learning research today.
\end{abstract}

\section{Introduction}

To start, we must be clear that by \textit{reproducibility} we are referring to the ability of an independent team to recreate the same qualitative results, and by \textit{replication} we are referring to the use of code to re-create the same results. These terms have been used inconsistently across different fields of study at various points throughout time ~\citep{Plesser2018}. Many major machine learning conferences have appointed reproducibility chairs, and in doing so have almost uniformly converged on using check-boxes to indicate that a submission includes code, or asking authors to answer vague questions about reproducibility. Some venues explicitly ask for code, others do not. Reviewers often believe that code indicates reproducibility. There appears to be prevailing belief that, if authors open-source their code and ensure the code reproduces the paper's results, we can solve the reproducibility crisis ~\citep{Forde2018,soton403913,Zaharia2018AcceleratingTM,Forde2018ReproducingML,Paganini2020,Gardner2018}. Our contention is that open source and associated replicability aides, are good --- but that this idealized notion is not a Pareto optimal improvement over papers that do not share source code. We argue that there are pros and cons to including source code with papers when we consider the long-term health of the field. The pros are widely known, and have been explored since the advent of digital communication \citep{Claerbout1992}. In this opinion piece, we argue the cons: current evidence (though more is dearly needed) suggests open-source code may improve replication, but creates new issues in reproducibility. 

Toward our argument, we have a fundamental axiom: if work can be replicated (i.e., using author's original code and data) but not reproduced, then the work constitutes, at best, ineffective science ~\citep{Drummond2009}. It is fine for authors to produce such works, but in the long term, we do not truly understand the mechanism of action or the truth of our methods unless they are reproducible. Ideally, we desire that the fraction of works that are reproducible increases over time. 

We will begin our argument in \autoref{sec:history} by noting prior history of reproducible research in other fields, and describing how we are slowly re-learning lessons discovered long ago that show how having code does not solve reproducibility by our axiom. We provide notable examples on how code did not benefit, or even delayed, important understanding of machine learning methods in \autoref{sec:code_harm} with the seminal word2vec and Adam. These are not arguments that these works are useless or ``wrong'', but that code negatively impacted better scientific understanding in the former, and provided no benefit in the latter. Finally we will conclude with the argument that conferences must create reproducibility tracks that include explicit guidelines for reviewers on how to judge submissions, so that we can advance the study of reproducibility before blindly stepping toward ineffective solutions. 

\section{We Have Forgotten History, Now We Are Repeating It} \label{sec:history}

The machine learning community has only just begun to expend serious effort towards the study of reproducibility with respect to itself as a domain. The discoveries are unnerving, and have strong parallels with historical findings in other domains. Significant early work in the study of code reproducibility was done by \cite{Hatton1993,Hatton1997,Hatton1994}, performing static analysis across C and FORTRAN code as well as having multiple implementations of the same algorithm, and providing the exact same inputs and parameters to each independent implementation. Their results found a high defect rate, more than 1 issue per 150 lines of code, and that the precision of independent implementations was only one significant figure. FORTAN and C still form the foundation of scientific computing, including machine learning packages like NumPy~\citep{Harris2020}, Tensorflow~\citep{Abadi2016}, and Pytorch~\citep{NIPS2019_9015}. These projects are important components of the computational foundation of our field, yet often focus on the pursuit of optimal performance at the expense of other goals such as maintainability and portability, the need for multi-disciplinary teams for success (e.g., where we often consider ``applications'' a secondary track or area that is often stigmatized), and most importantly, the high difficulty of verifying the correctness of the equations and math implemented \citep{Carver2007}. Indeed as history repeats itself, recent work has identified cases where the same models implemented with different packages or hardware accelerators present reduced precision or accuracy to the order of one significant figure and even greater variation in run-time consistency \citep{Zhuang2021}. Even within a single set of hardware and implementation, our most widely used libraries often have non-deterministic implementations that can cause 10\% variances in results \citep{10.1145/3324884.3416545}.

A more nuanced version of the above point stems from how we define replication: does it simply involve the code, or must it also include the data? The latter is how the terminology is most historically used, and common in other sciences~\citep{Plesser2018}. This is challenging in machine learning due to the intrinsic ``fuzzyness'' of what we are working toward: we intrinsically wish to use machine learning for tasks where thorough specification of the data is too difficult to implement in code. We can again look to other fields, like software engineering, that attempted to perform reproductions that included the data process over software repositories \citep{10.1007/s10664-011-9181-9}. Their work found that missing or minute details could prevent or significantly impede reproduction. Indeed it becomes unsurprising then that we have only recently discovered considerable labeling issues within foundational datasets like MNIST, CIFAR, and ImageNet \citep{Northcutt2021}. While data sheets and model cards have been proposed to partially address this issue, \citep{Gebru2021,10.1145/3287560.3287596} they are proposed without any scientific study to answer if these interventions mitigate the underlying problem. It is good for producers and users of datasets to carefully think about the data in use, but we fear that absent evidence, these approaches may have no direct tangible impact\footnote{Their ability to change thoughts and focus areas of others, creating positive secondary impacts, is more likely, but a separate matter beyond our discussion. }. Indeed studies of dataset replication (where no model card exists) have been shockingly successful in some ways (no evidence of adaptive over-fitting) and identify concerns not addressed in model cards or data sheets \citep{pmlr-v119-engstrom20a} with similar results over applied domains such as social media analysis \citep{Geiger2021}. 

As such, we argue that there is extensive prior evidence that predicts the current trends in machine learning reproducible research: having code available means relatively little to the question of reproducibility, especially in light of inconsistent methods of comparison used through decades of machine learning literature, leading to invalid conclusions of ``improvement'' \citep{Bradley1997,Alpaydin:1999:CTC:339993.339999,Cawley:2010:OMS:1756006.1859921,Demsar:2006:SCC:1248547.1248548,JMLR:v17:benavoli16a,Bouthillier2021,pmlr-v97-bouthillier19a,Dror2017,dror-etal-2018-hitchhikers,dror-etal-2019-deep}, necessitating that even a system with no bit-rot would not solve the concerns of our field. 

\section{How Can Open Sourced Code Harm us?} \label{sec:code_harm}

Given that we have clear evidence that simply having original source code is not sufficient to enable reproducibility, we must now ask: can withholding code ever lead to an improvement in reproducibility? It is important to be clear that we are not arguing that no-code is always or even \textit{usually} better. We are arguing that a lack of code creates a different kind of forcing function for adoption. We recognize\footnote{Without the same quality of evidence, indeed we are not sure how to design a good experiment for this. But this is an opinion paper, so we feel some indignant right to be opinionated.} that code sharing is likely to lead to faster adoption of a method that works, but obscures long-term benefits to reproducible work. If a paper's method must be re-implemented due to a lack of code, this process organically validates said method. The paper's method only gets used and cited\footnote{There are certainly edge cases where a method that does not replicate will be used and cited, we are talking in the more general broader case of directly building upon or relying on a method. }  when others can successfully reproduce it, converging on methods that work and forcing deeper understanding by a broader population. Further still, this forces the community at large to be effective communicators and to better understand the details and science required to reproduce one's results. The need to enable reproducibility drove \cite{Taswell1998} to develop a proposal to better specify wavelet transforms, which also enabled better replicability of his methods. We find tangential evidence for this within machine learning where 36\% of papers could not be reproduced from their content, even though many provided source code \citep{Raff2019_quantify_repro}. 
To further exemplify how we believe this to be an issue, we will draw from highly successful academics to critique with a bias to avoid undue harm or stress to early career researchers (in similar spirit to \cite{Lipton2019} ). 

The seminal word2vec \citep{Mikolov2013a} algorithm is our first consideration. A publication who's ubiquity and impact in research and application is enormous, and to the best of our knowledge, has never been replicated. Understanding how and why word2vec worked was studied by many ~\citep{Goldberg2014} due to its utility and effectiveness, but was done through the originally released code (or direct translations into other languages). Yet it took six years for any public documentation of the fact that the paper and code simply do not perform the same steps~\cite{Bhat2019}, making it impossible for anyone to reproduce. 

Clearly, word2vec was important and valuable for the community, but there are counterfactual questions that we argue suggest the long-term health of our research would have been better if \cite{Mikolov2013a} never released their source code. First, there is an unknown amount of person hours wasted by researchers, students, and others attempting to understand the mechanisms of an algorithm that was inhibited by faulty foundations\footnote{Not to mention feelings of inadequacy, anxiety, and stress by students attempting to become researchers in what is already a needlessly high-stress environment.}. Second, failure to reproduce by others would be a forcing function on the original authors to re-examine their code and paper to correctly document how and why it works. By releasing the code, this feedback cycle is inhibited. This could also explain how follow-up work with paragraph2vec \citep{pmlr-v32-le14} has similarly evaded reproduction, even by the paper's co-authors\footnote{\url{https://groups.google.com/g/word2vec-toolkit/c/Q49FIrNOQRo/m/DoRuBoVNFb0J}}.

A different perspective on this matter is seen in the Adam optimizer \citep{Kingma2015}, which has become a widely used default method. This case is interesting in that the simplicity of the approach has enabled many reproductions, but both the code and the paper lack details on how the default parameter values were derived. Subtle corrections to the math of Adam in weight decay \citep{Loshchilov2019} and the $\epsilon$ parameter \citep{Yuana} can yield large improvements in the quality of results, as the default values of Adam are not ideal for all cases. While we should, in general, have no reason to believe in a one-size-fits-all approach, the lack of study around these details is itself lending to reproducibility challenges in our field: the ``right'' way to set these parameters (amongst dozens of others in a network) was unstudied, and many sub-fields began tweaking the defaults for their kinds of networks, creating confusion and slowing reproduction of subsequent research. This kind of issue is not new. Poorly documented accounts of differences in LBFGS~\citep{Liu1989} implementation results can be found~\footnote{\url{https://discourse.julialang.org/t/optim-jl-vs-scipy-optimize-once-again/61661/5?page=2}}, though we are not aware of any thorough documentation or study of them. This again suggests an issue with an incomplete description in the paper, a problem that code can not reduce --- but can hide for a period of time. 

We also argue that relying on open source code creates an academic moral hazard. Distilling the the essence of the scientific contribution, and communicating it effectively, is the task of an author. Although code does not solve reproducibility, it does enable replication and provides short term benefits in citation rate and adoption~\citep{Raff2022}, thus allows the manuscript to defer ``nuscance'' details to the code. Reviewers can run the code to confirm that ``it works'', without checking that the code actually performs the method described precisely or simply be unaware of key confounding design choices. We preemptively rebut an argument that having code does allow checking an approach. We rebuke this argument by noting that decades of research shows that reading and debugging code does not ensure the same kind of mental processing as reading prose ~\citep{Ivanova2020,Perkins1988,Letovsky1987}. This is confirmed by the demonstrated positive impact of high quality code comments on understanding code~\citep{Nurvitadhi}. As such the reading of code is a more challenging mental process than reading well constructed prose in a paper, and while helpful is not an alternative to effective communication.  This fact, combined with the examples of Adam and word2vec show ways that code, regardless of how easy it is to implement, can harm reproducibility. We fear that an over-emphasis on code will seed new reproducibility problems.
\section{Conclusion} \label{sec:conclusion} \label{sec:conclusion}

We remind the reader that we are not arguing that open sourcing code is bad. Open source code is valuable, but is not a panacea for reproducible research. A lack of advances in the science of reproducible research will lead to negative long-term artifacts that are expensive to remedy. Almost all current conference efforts focus solely around open-source. We argue there is strong evidence that this will not improve or solve the problem of reproducibility. In our minds, the primary issue is that, similar to other fields of science, the study of better reproducibility practices will have a cost that is not rewarded ~\citep{Poldrack2019}. We call upon the conference chairs at our primary publication venues create those incentives with two simple changes.

First and foremost, specialized tracks on reproducibility, and rewards for reproducibility, should become standard at all conferences. The Conference of the Association for Computing Machinery Special Interest Group in Information Retrieval (SIGIR) is the only conference we are aware of that has taken a positive step toward the underlying issue by creating a reproducibility track\footnote{\url{https://sigir.org/sigir2022/call-for-reproducibility-track-papers/}}. However we argue that SIGIR's current scope for the track is too small: it only considers papers revisiting prior techniques and expanding their study (e.g., does this method work on additional problems, or under altered conditions/constraints?). Reproducibility tracks should encourage research into the questions of reproducibility itself: user studies, incentive structures, and generally accept a broader scope of work. This is necessary because the science of quantified and well-studied reproducibility receives little attention across all fields of study. This is also a need as we observe key work revisiting fundamental foundations of our field yet appearing only on arXiv \citep{Recht2019,Recht2018,Ahn2022} or outside of our top publishing venues \citep{Barz2019}. By creating reproducibility tracks with a broader scope, we can immediately create stronger incentives for this needed work. 

Second, and related to the prior concern, is the lack of guidelines given to reviewers for evaluating reproducibility. If we do not state explicitly what is and is-not acceptable, we only increase the noise floor of acceptance and of reproducible work. 
Of primary concern is that we rely too heavily on intuitions about what will improve reproducibility because we lack well-defined measurements. We must instruct reviewers to use a standard that requires quantification, even if imperfect, to elevate the literature from hopes to science. For example, \citet{Raff2020c} uses imperfect data with censored labels to study time to implement an algorithm, but adjusts the model to address data limitations. Anecdotally, we report as a reviewer in the first NeurIPS Datasets and Benchmarks track, multiple reviewers complain of a lack of novel algorithms, for a track that is explicitly not about algorithms, and for which no reviewer guidelines about what qualities or desiderata should be included in an accepted paper for the new track. Similarly we point to \cite{Tran2020} which showed bias in the acceptance process of OpenReview ICLR papers, with ``\textit{The main argument for rejection is the the analysis done in the paper is not typical of ICLR research}''\footnote{\url{https://openreview.net/forum?id=Cn706AbJaKW}}. We consider these cases tragic in that no explicit instructions appear to exist both around replication, self-study as a field, or the ability to accept work so novel that it does not fit our existing mold. Reproducible research need not be ``novel'' in method, require proofs, or advanced math --- its criteria should be quantified evidence toward any aspect of how (non)reproducible work gets accepted, encouraged, propagated, discovered, and any other aspect that would reasonably relate to the question of reproducibility. If we can't accept quantified criticism of our field and institutions, we are lost as a scientific discipline. 

\bibliography{references}
\bibliographystyle{iclr2022_conference}

\end{document}